\documentclass[11pt,a4paper]{article}
\usepackage[nohyperref]{acl2018}
\usepackage{times}
\usepackage{latexsym}
\usepackage{url}
\usepackage{graphicx}
\usepackage{amssymb}
\usepackage{amsmath}
\usepackage{multirow}
\usepackage{graphicx}
\usepackage{multirow}
\usepackage{algorithm}
\usepackage{algorithmic}
\usepackage{amssymb}
\usepackage{amsmath}
\usepackage{dsfont}
\aclfinalcopy 

\title{Semantic Parsing with Syntax- and Table-Aware SQL Generation}

\author{Yibo Sun$^\S$\thanks{\ \ Work is done during internship at Microsoft Research Asia.}\ \ , Duyu Tang$^\ddag$, Nan Duan$^\ddag$, Jianshu Ji$^\natural$, Guihong Cao$^\natural$,  \\
	\bf Xiaocheng Feng$^\S$, Bing Qin$^\S$, Ting Liu$^\S$ Ming Zhou$^\ddag$ \\
	$^\S$Harbin Institute of Technology, Harbin, China\\
	$^\ddag$Microsoft Research Asia, Beijing, China \\
	$^\natural$Microsoft AI and Research, Redmond WA, USA\\
	{\small \tt \{ybsun,xcfeng,qinb,tliu\}@ir.hit.edu.cn}\\
	{\small \tt \{dutang,nanduan,jianshuj,gucao,mingzhou\}@microsoft.com}}

\date{}

\begin{document}
	\maketitle
	
\begin{abstract}
We present a generative model to map natural language questions into SQL queries.
Existing neural network based approaches typically generate a SQL query word-by-word, however, a large portion of the generated results are incorrect or not executable due to the mismatch between question words and table contents.
Our approach addresses this problem by considering the structure of table and the syntax of SQL language.
The quality of the generated SQL query is significantly improved through (1) learning to replicate content from column names, cells or SQL keywords;
and (2) improving the generation of WHERE clause by leveraging the column-cell relation.
Experiments are conducted on WikiSQL, a recently released dataset with the largest question-SQL pairs.
Our approach significantly improves the state-of-the-art execution accuracy from 69.0\% to 74.4\%.
\end{abstract}


\section{Introduction}

We focus on semantic parsing that maps natural language utterances to executable \mbox{programs}
 \cite{zelle1996learning,wong2007learning,zettlemoyer2007online,kwiatkowski2011lexical,pasupat-liang:2015:ACL-IJCNLP,iyer-EtAl:2017:Long,iyyer-yih-chang:2017:Long}.
In this work, we regard SQL as the program language, which could be executed on a table or a relational database to obtain an outcome.
Datasets are the main driver of progress for statistical approaches in semantic parsing \cite{liang2016learning}.
Recently, \newcite{zhong2017seq2sql} release WikiSQL, the largest hand-annotated semantic parsing dataset which is an order of magnitude larger than other datasets in terms of both the number of logical forms and the number of tables.
Pointer network \cite{vinyals2015pointer} based approach is developed, which generates a SQL query word-by-word through replicating from a word sequence consisting of question words, column names and SQL keywords.
However, a large portion of generated results are incorrect or not executable due to the mismatch between question words and column names (or cells).
This also reflects the real scenario where users do not always use exactly the same column name or cell content to express the question.

To address the aforementioned problem, we present a generative semantic parser that considers the structure of table and the syntax of SQL language.
The approach is partly inspired by the success of structure/grammar driven neural network approaches in semantic parsing \cite{xiao-dymetman-gardent:2016:P16-1,krishnamurthy-dasigi-gardner:2017:EMNLP2017}.
Our approach is based on pointer networks, which encodes the question into continuous vectors, and synthesizes the SQL query
with three channels. The model learns when to generate a column name, a cell or a SQL keyword.
%
We further incorporate column-cell relation 
to mitigate the ill-formed outcomes.

\begin{figure*}[t]
	\centering
	\includegraphics[width=0.75\textwidth]{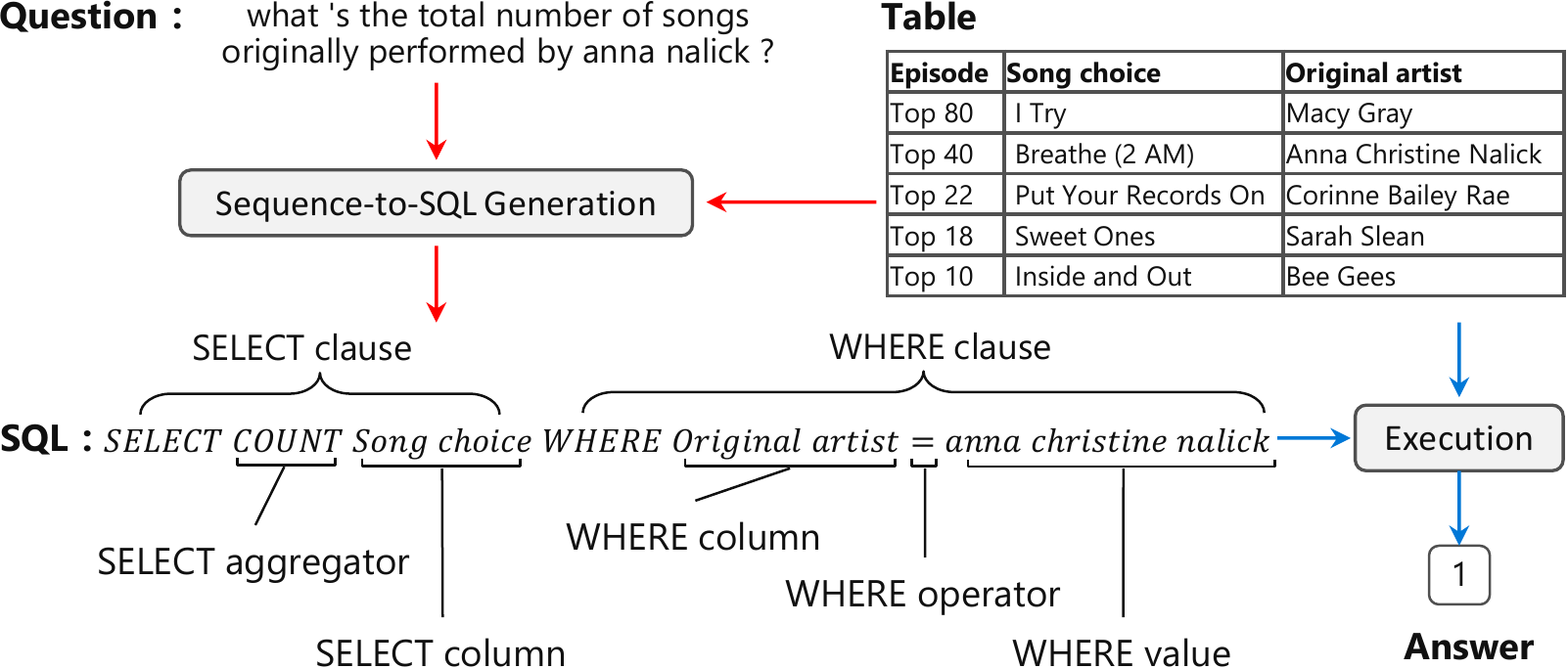}	\caption{An brief illustration of the task. The focus of this work is sequence-to-SQL generation.}
	\label{fig:task}
\end{figure*}
We conduct experiments on WikiSQL.
Results show that our approach outperforms existing systems,
 improving state-of-the-art execution accuracy to 74.4\%
and logical form accuracy to 60.7\%.
Extensive analysis reveals the advantages and limitations of our approach.




\section{Task Formulation and Dataset}
As shown in Figure \ref{fig:task}, we 
focus on sequence-to-SQL generation in this work. 
Formally, the task takes a question $q$ and a table $t$ consisting of $n$ column names and $n \times m$ cells as the input, and outputs a SQL query $y$.
We do not consider the join operation over multiple relational tables, which we leave in the future work.

We use WikiSQL \cite{zhong2017seq2sql}, the largest hand-annotated semantic parsing dataset to date which 
consists of 87,726 questions and SQL queries distributed across 26,375 tables from Wikipedia.



\section{Related Work}
\paragraph{Semantic Parsing.}
Semantic parsing
aims to map natural language utterances to programs (e.g., logical forms), which will be executed to obtain the answer (denotation) \cite{Zettlemoyer05,liang2011learning,berant2013semantic,poon2013grounded,krishnamurthy2013jointly,pasupat-liang:2016:P16-1,sun2016table,jia-liang:2016:P16-1,kovcisky-EtAl:2016:EMNLP2016,LinWPVZE2017:TR}.
Existing studies differ from
(1) the form of the knowledge base, e.g. facts from Freebase, a table (or relational database), an image \cite{suhr-EtAl:2017:Short,johnson2017inferring,hu2017learning,DBLP:journals/corr/abs-1711-05240} or a world state \cite{long-pasupat-liang:2016:P16-1};
(2) the program language, e.g. first-order logic, lambda calculus, lambda DCS, SQL, parameterized neural programmer \cite{yin2015neural,neelakantan2016learning}, or coupled distributed and symbolic executors \cite{mou17};
(3) the supervision used for learning the semantic parser, e.g. question-denotation pairs, binary correct/incorrect feedback \cite{artzi2013weakly}, or richer supervision of question-logical form pairs \cite{dong-lapata:2016:P16-1}.
In this work, we study semantic parsing over tables,
which is critical for users to access relational databases with natural language, and could serve users' information need for structured data on the web.
We use SQL as the program language, which has a broad acceptance to programmers.

\paragraph{Natural Language Interface for Databases.}
Our work relates to the area of accessing database with natural language interface \cite{dahl1994expanding,brad2017dataset}.
\newcite{popescu2003towards} use a parser to parse the question, and then use lexicon matching between question and the table column names/cells.
\newcite{giordani2012translating} parse the question with dependency parser, compose candidate SQL queries with heuristic rules, and use kernel based SVM ranker to rank the results.
\newcite{li2014constructing} translate natural language utterances into SQL queries based on dependency parsing results, and interact with users to ensure the correctness of the interpretation process.
\newcite{yaghmazadeh2017type} build a semantic parser on the top of SEMPRE \cite{pasupat-liang:2015:ACL-IJCNLP} to get a SQL sketch, which only has the SQL shape and will be subsequently completed based on the table content.
\newcite{iyer-EtAl:2017:Long} maps utterances to SQL queries through sequence-to-sequence learning.
User feedbacks are incorporated to reduce the number of queries to be labeled.
\newcite{zhong2017seq2sql} develop an augmented pointer network, which is further improved with reinforcement learning for SQL sequence prediction.
\newcite{xu2017sqlnet} adopts sequence-to-set model to predict WHERE columns, and uses attentional model to predict the slots in where clause.

Different from \cite{iyer-EtAl:2017:Long,zhong2017seq2sql}, our approach leverages SQL syntax and table structure.
Compared to \cite{popescu2003towards,giordani2012translating,yaghmazadeh2017type}, our approach is end-to-end learning and independent of syntactic parser or manually designed templates.
We are aware of existing studies that combine reinforcement learning and maximum likelihood estimation (MLE) \cite{guu-EtAl:2017:Long,mou17,liang-EtAl:2017:Long}. However, the focus of this work is the design of the neural architecture, despite we also implement a RL strategy (refer to \S \ref{section:improved-rl}).

\paragraph{Structure/Grammar Guided Neural Decoder}
Our approach could be viewed as an extension of the sequence-to-sequence learning \cite{sutskever2014sequence,Bahdanau2015} with a tailored neural decoder driven by the characteristic of the target language \cite{yin2017syntactic,rabinovich2017abstract}.
Methods with similar intuitions have been developed for language modeling \cite{dyer-EtAl:2016:N16-1}, neural machine translation \cite{wu-EtAl:2017:Long2} and lambda calculus based semantic parsing  \cite{dong-lapata:2016:P16-1,krishnamurthy-dasigi-gardner:2017:EMNLP2017}.
The difference is that our model is developed for sequence-to-SQL generation, in which table structure and SQL syntax are considered.

\begin{figure*}[t]
	\centering
	\includegraphics[width=\textwidth]{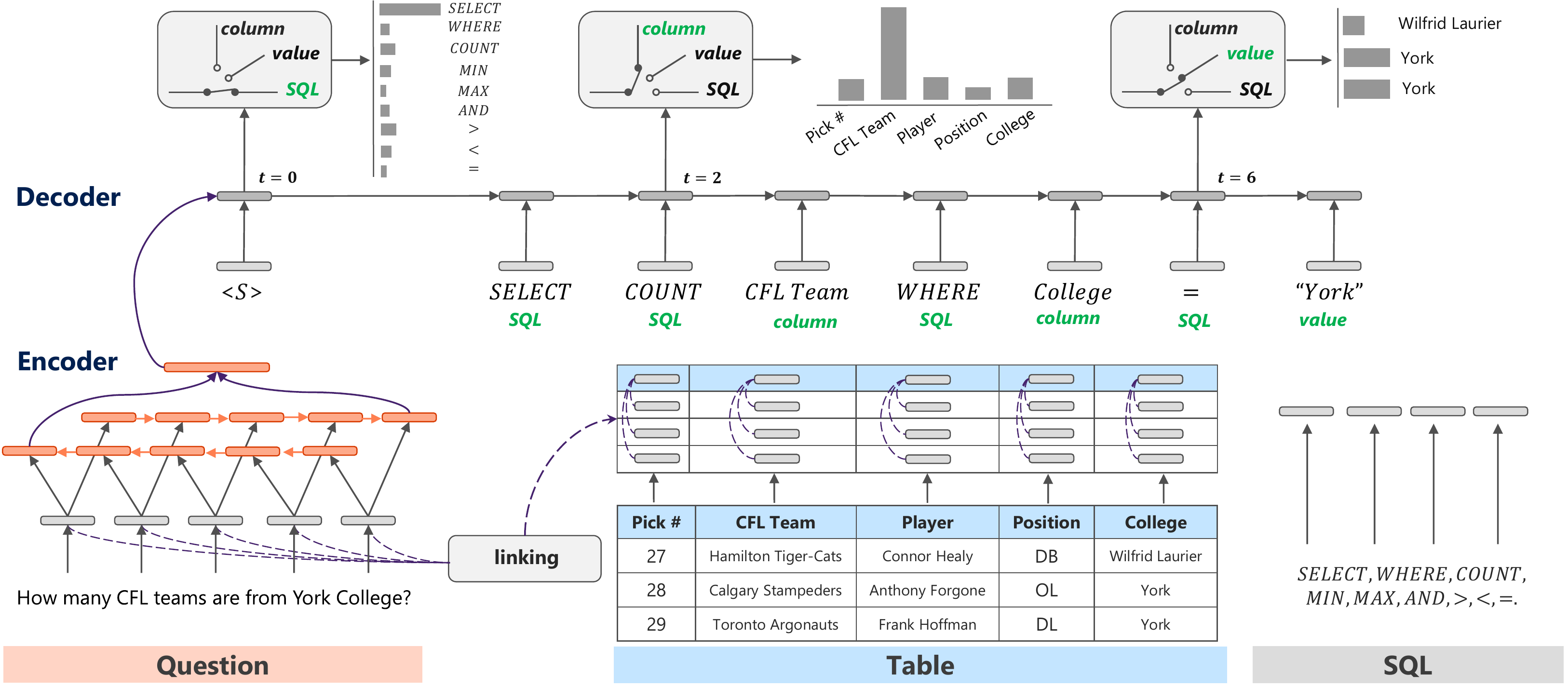}
	\caption{An illustration of the proposed approach. At each time step, a switching gate selects a channel to predict a column name (maybe composed of multiple words), a cell or a SQL keyword. The words in green below the SQL tokens stand for the results of the switching gate at each time step.}
	\label{fig:model}
\end{figure*}

\section{Methodology}
We first describe the background on pointer networks, 
and then present our approach that considers the table structure and the SQL syntax.

\subsection{Background: Pointer Networks}
Pointer networks is originally introduced by \cite{vinyals2015pointer}, which takes a sequence of elements as the input and outputs a sequence of discrete tokens corresponding to positions in the input sequence.
The approach has been successfully applied in reading comprehension \cite{kadlec-EtAl:2016:P16-1} for pointing to the positions of answer span from the document, and also in sequence-to-sequence based machine translation \cite{gulcehre-EtAl:2016:P16-1} and text summarization \cite{gu-EtAl:2016:P16-1} for replicating rare words from the source sequence to the target sequence.

The approach of \newcite{zhong2017seq2sql} is based on pointer networks.
The encoder is a \mbox{recurrent} neural network (RNN) with gated recurrent unit (GRU) \cite{cho-EtAl:2014:EMNLP2014}, whose input is the concatenation of question words, words from column names and SQL keywords.
The decoder is another GRU based RNN, which works in a sequential way and generates a word at each time step.
The generation of a word is actually selectively replicating a word from the input sequence, the probability distribution
of which is calculated with an attention mechanism \cite{Bahdanau2015}.
The probability of generating the $i$-th word $x_i$ in the input sequence at the $t$-th time step is calculated as Equation \ref{equa:pnt}, where $h^{dec}_t$ is the decoder hidden state at the $t$-th time step, $h^{enc}_i$ is the encoder hidden state of the word $x_i$, $W_a$ is the model parameter.
\begin{equation}\label{equa:pnt}
p(y_t=x_i| y_{<t}, x) \propto exp(W_a [h^{dec}_t ; h^{enc}_i])
\end{equation}
It is worth to note that if a column name consists of multiple words (such as ``original artist'' in Figure \ref{fig:task}), these words are separated in the input sequence.
The approach has no guarantee that a multi-word column name could be successively generated,
which would affect the executability of the generated SQL query.



\subsection{{STAMP}: Syntax- and Table- Aware seMantic Parser }
\label{sec:framework}
Figure~\ref{fig:model} illustrates an overview of the proposed model, which is abbreviated as STAMP.
Different from \newcite{zhong2017seq2sql}, word is not the basic unit to be generated in STAMP.
As is shown, there are three ``channels'' in STAMP, among which the column channel predicts a column name, the value channel predicts a table cell and the SQL channel predicts a SQL keyword.
Accordingly, the probability of generating a target token is formulated in Equation \ref{equa:our}, where $z_t$ stands for the channel selected by the switching gate, $p_z(\cdot)$ is the probability to choose a channel, and $p_w(\cdot)$ is similar to Equation \ref{equa:pnt} which is a probability distribution over the tokens from one of the three channels.
\begin{equation}\label{equa:our}
p(y_t| y_{<t}, x) = \sum_{z_t} p_w(y_t | z_t, y_{<t}, x) p_z(z_t| y_{<t}, x)
\end{equation}
One advantage of this architecture is that it inherently addresses the problem of generating partial column name/cell because an entire column name/cell is the basic unit to be generated.
Another advantage is that the column-cell relation and question-cell connection
can be naturally integrated in the model, which will be described below.

Specifically, our encoder takes a question as the input. Bidirectional RNN with GRU unit is applied to the question, and the concatenation of both ends is used as the initial state of the decoder.
Another bidirectional RNN is used to compute the representation of a column name (or a cell), in case that each unit contains multiple words  \cite{dong2015hybrid}.
Essentially, each channel is an attentional neural network.
For cell and SQL channels, the input of the attention module only contains the decoder hidden state and the representation of the token to be calculated as follows,
\begin{equation}
p_w^{sql}(i)  \propto exp(W_{sql} [h^{dec}_t;e^{sql}_i])
\end{equation}
where $e^{sql}_i$ stands for the representation of the $i$-th SQL keyword.
As suggested by \cite{zhong2017seq2sql}, we also concatenate the question representation into the input of the column channel in order to improve the accuracy of the SELECT column.
We implement the switching gate with a feed-forward neural network, in which the output is a $softmax$ function and the input is the decoder hidden state $h^{dec}_t$.
%

\subsection{Improved with Column-Cell Relation}
\label{section:improved-column-cell}
We further improve the STAMP model by considering the column-cell relation,
which is important for predicting the WHERE clause.

On one hand, the column-cell relation could improve the prediction of SELECT column.
We observe that a cell or a part of it typically appears at the question acting as the WHERE value, such as \textit{``anna nalick''} for ``\textit{anna christine nalick}'').
However, a column name might be represented with a totally different utterance, which is a ``semantic gap''.
Supposing the question is ``How many schools did player number 3 play at?'' and the SQL query is ``Select count School\\Club Team where No. = 3''. We could see that the column names ``School\\Club Team'' and ``No.'' are different from their corresponding utterances ``schools'', ``number'' in natural language question. 
Thus, table cells could be regarded as the pivot that connects the question and column names (the ``linking'' component in Figure \ref{fig:model}).
For instance, taking the question from Figure \ref{fig:model},
the word ``\textit{York}'' would help to predict the column name as ``\textit{College}'' rather than ``\textit{Player}''.
There might be different possible ways to implement this intuition.
We use cell information to enhance the \mbox{column} name representation in this work.
The vector of a column name is further concatenated with a question-aware cell vector, which is weighted averaged over
the cell vectors belonging to the same column. The probability distribution in the column channel is calculated as Equation \ref{equa:improved-column-channel}.
We use the number of cell words occurring in the question to measure
the importance of a cell, which is further normalized through a $softmax$ function to yield the final weight $\alpha^{cell}_j \in [0,1]$. An alternative measurement is to using an additional attention model whose input contains the question vector and the cell vector. We favor to the intuitive and efficient way in this work.
\begin{equation}\label{equa:improved-column-channel}
p_w^{col}(i)  \propto exp(W_{col} [h^{dec}_t;h^{col}_i;\sum_{j \in col_i} \alpha^{cell}_j h^{cell}_j])
\end{equation}


On the other hand, the column-cell relation could improve the prediction of the WHERE value.
To yield an executable SQL, the WHERE value should be a cell that belongs to the same WHERE column\footnote{This constraint is suitable in this work as we do not consider the nested query in the where clause, such as ``\textit{where College = select College from table}'', which is also the case not included in the WikiSQL dataset. We leave generating nested SQL query in the future work.}.
Taking Figure \ref{fig:model} as an example, it should be avoided to predict a where clause like ``\textit{Player = York}'' because the cell ``\textit{York}'' does not belong to the column name ``\textit{Player}''.
To achieve this, we incorporate an global variable to memorize the last predicted column name. When the switching gate selects the value channel, the \mbox{cell} distribution is only calculated over the cells belonging to the last predicted column name.
Furthermore, we incorporate an additional probability distribution over cells based on the aforementioned word co-occurrence between the question and cells, and weighted average two cell distributions, which is calculated as follows.
\begin{equation}
p_w^{cell}(j) = \lambda \hat{p}_w^{cell}(j) + (1-\lambda) \alpha^{cell}_j
\end{equation}
where $\hat{p}_w^{cell}(j)$ is the standard probability distribution obtained from the attentional neural network, and $\lambda$ is a hyper parameter which is tuned on the dev set.


\subsection{Improved with Policy Gradient}
\label{section:improved-rl}
The model described so far could be conventionally learned via cross-entropy loss over question-SQL pairs.
However, different SQL queries might be executed to yield the same result, and possible SQL queries of different variations
could not be exhaustively covered in the training dataset.
Two possible ways to handle this are (1) shuffling the WHERE clause to generate more SQL queries, and (2) using 
reinforcement learning (\mbox{RL}) which 
%
regards the correctness of the executed output as the goodness (reward) of the generated SQL query.
We follow \newcite{zhong2017seq2sql} and adopt a policy gradient based approach.
We use a baseline strategy \cite{zaremba2015reinforcement} to decrease the learning variance.
The expected reward \cite{williams1992simple} for an instance is calculated as $\mathop{\mathbb{E}}(y_g)=\sum_{j=1}^{k}logp(y_j) R(y_j, y^g)$, where $y^g$ is the ground truth SQL query, $y_j$ is a generated SQL query, $p(y_j)$ is the probability of $y_j$ being generated by our model, and $k$ is the number of sampled SQL queries.
$R(y_j, y^g)$ is the same reward function defined by \newcite{zhong2017seq2sql}, which is
$+1$ if $y_j$ is executed to yield the correct answer;
$-1$ if $y_j$ is a valid SQL query and is executed to get an incorrect answer;
and $-2$ if $y_j$ is not a valid SQL query.
In this way, model parameters could be updated
with policy gradient over question-answer pairs.

\subsection{Training and Inference}
As the WikiSQL data contains rich supervision of question-SQL pairs, we use them to train model parameters.
The model has two cross-entropy loss functions, as given below.
One is for the switching gate classifier ($p_z$) and another is for the attentional probability distribution of a channel ($p_w$). 
\begin{equation}
l = -\sum_t log p_z(z_t| y_{<t}, x)-\sum_t log p_w(y_t | z_t, y_{<t}, x)
\end{equation}
Our parameter setting strictly follows \newcite{zhong2017seq2sql}.
We represent each word using word embedding\footnote{\url{http://nlp.stanford.edu/data/glove.840B.300d.zip}} \cite{pennington-socher-manning:2014:EMNLP2014} and the mean of the sub-word embeddings of all the n-grams in the word  \cite{hashimoto2016joint}\footnote{\url{http://www.logos.t.u-tokyo.ac.jp/~hassy/publications/arxiv2016jmt/jmt_pre-trained_embeddings.tar.gz}}.
The dimension of the concatenated word embedding is 400.
We clamp the embedding values to avoid over-fitting.
We set the dimension of encoder and decoder hidden state as 200.
During training, we randomize model parameters from a uniform distribution with fan-in and fan-out, set batch size as 64, set the learning rate of SGD as 0.5, and update the model with stochastic gradient decent.
Greedy search is used in the inference process.
We use the model trained from question-SQL pairs as initialization and use RL strategy to fine-tune the model.
SQL queries used for training RL are sampled based on the probability distribution of the model learned from question-SQL pairs.
We tune the best model on the dev set and do inference on the test set for only once.
This protocol is used in model comparison as well as in ablations.

\begin{table*}[t]
	\centering
	\begin{tabular}{l|cc|cc}
		\hline
		\multirow{2}{*}{{Methods}} & \multicolumn{2}{c|}{Dev} & \multicolumn{2}{c}{Test} \\
		\cline{2-5}
		& {Acc$_{lf}$} & {Acc$_{ex}$} & {Acc$_{lf}$} & {Acc$_{ex}$}\\
		\hline
		Attentional Seq2Seq & 23.3\%& 37.0\%& 23.4\%& 35.9\% \\
		Aug.PntNet~\cite{zhong2017seq2sql} & 44.1\%& 53.8\%& 43.3\%& 53.3\% \\
		Aug.PntNet (re-implemented by us) &51.5\%& 58.9\%& 52.1\%& 59.2\% \\
		Seq2SQL (no RL)~\cite{zhong2017seq2sql} &  48.2\%& 58.1\%& 47.4\% &57.1\% \\
		Seq2SQL~\cite{zhong2017seq2sql} & 49.5\%& 60.8\%& 48.3\%& 59.4\%\\
		SQLNet~\cite{xu2017sqlnet} & -- & 69.8\% & -- & 68.0\%\\
		\newcite{DBLP:journals/corr/abs-1801-00076} & -- & 71.1\%& -- & 69.0\%\\
		\hline
		
		STAMP (w/o cell)& 58.6\%& 67.8\%& 58.0\%& 67.4\%\\
		STAMP (w/o column-cell relation) & 59.3\%& 71.8\%& 58.4\%& 70.6\%\\
		STAMP  & 61.5\%& 74.8\%& 60.7\%& 74.4\% \\
		STAMP+RL& 61.7\%&  75.1\%& 61.0\%&  74.6\%\\
		\hline
	\end{tabular}
	\caption{Performances of different approaches on the WikiSQL dataset. Two evaluation metrics are logical form accuracy (Acc$_{lf}$) and execution accuracy (Acc$_{ex}$).
		Our model is abbreviated as (\textbf{STAMP}).}
	\label{table:compare-to-other-alg}
\end{table*}
\section{Experiment}
We conduct experiments on the WikiSQL dataset\footnote{\url{https://github.com/salesforce/WikiSQL}}, which includes $61,297/9,145/17,284$ examples in the training/dev/test sets.
Each instance consists of a question, a table, a SQL query and a result.
Following \newcite{zhong2017seq2sql}, we use two evaluation metrics. One metric is logical form accuracy (Acc$_{lf}$), which measures the percentage of the generated SQL queries that have exact string match with the ground truth SQL queries.
Since different SQL queries might obtain the same result, another metric is
execution accuracy (Acc$_{ex}$), which measures the percentage of the generated SQL queries that obtain the correct answer.

\subsection{Model Comparisons}
After released, WikiSQL dataset has attracted a lot of attentions from both industry and research communities.
\newcite{zhong2017seq2sql} develop following methods, including (1)
	{Aug.PntNet} which is an end-to-end learning pointer network approach;
	(2) {Seq2SQL (no RL)}, in which two separate classifiers are trained for SELECT aggregator and SELECT column, separately; and (3) {Seq2SQL}, in which reinforcement learning is further used for model training.
	Results of the attentional sequence-to-sequence learning baseline ({Attentional Seq2Seq}) are also reported in \cite{zhong2017seq2sql}. 
	\newcite{xu2017sqlnet} develop {SQLNet}, which predicts SELECT clause and WHERE clause separately. Sequence-to-set neural architecture and column attention are adopted to predict the WHERE clause.
Similarly, \newcite{DBLP:journals/corr/abs-1801-00076} develop tailored modules to handle three components of SQL queries, respectively.


Our model is abbreviated as (\textbf{STAMP}), which is short for Syntax- and Table- Aware seMantic Parser.
The STAMP model in Table \ref{table:compare-to-other-alg} stands for the model we describe in \S \ref{sec:framework} plus \S \ref{section:improved-column-cell}.
\mbox{STAMP+RL} is the model that is fine-tuned with the reinforcement learning strategy as described in \S \ref{section:improved-rl}.
We implement a simplified version of our approach (w/o cell), in which WHERE values come from the question. Thus, this setting differs from Aug.PntNet in the generation of WHERE \mbox{column}.
We also study the influence of the relation-cell relation (w/o column-cell relation) through removing the enhanced column vector, which is calculated by weighted averaging cell vectors.

From Table \ref{table:compare-to-other-alg}, we can see that STAMP performs better than existing systems on WikiSQL.
Incorporating RL strategy does not significantly improve the performance.
Our simplified model, STAMP (w/o cell), achieves better accuracy than Aug.PntNet,
which further reveals the effects of the column channel.
Results also demonstrate the effects of incorporating the column-cell relation, removing which leads to about 4\% performance drop in terms of Acc$_{ex}$.

\subsection{Model Analysis: Fine-Grained Accuracy}
We analyze the STAMP model from different perspectives in this part.

\begin{table*}[t]
	\centering
	\begin{tabular}{l|ccc|ccc}
		\hline
		\multirow{2}{*}{Methods} & \multicolumn{3}{c|}{Dev} & \multicolumn{3}{c}{Test} \\
		\cline{2-7}
		& {Acc$_{sel}$} & {Acc$_{agg}$} & Acc$_{where}$ & {Acc$_{sel}$} & {Acc$_{agg}$}& Acc$_{where}$\\
		\hline
		Aug.PntNet (reimplemented by us) & 80.9\%& 89.3\%& 62.1\%& 81.3\%& 89.7\%& 62.1\%\\
		Seq2SQL~\cite{zhong2017seq2sql} & 89.6\%& 90.0\%& 62.1\%& 88.9\%& 90.1\%& 60.2\%\\
		SQLNet~\cite{xu2017sqlnet} & 91.5\% & 90.1\%& 74.1\%& 90.9\% & 90.3\%& 71.9\%\\
 		\newcite{DBLP:journals/corr/abs-1801-00076} & 92.5\% & 90.1\% & 74.7\%& 91.9\%& 90.3\%& 72.8\% \\
		
		\hline
		
		STAMP  (w/o cell)& 89.9\%& 89.2\%& 72.1\%& 89.2\%& 89.3\%& 71.2\%\\
		STAMP  (w/o column-cell relation) & 89.3\%& 89.2\%& 73.2\%& 88.8\%& 89.2\%& 71.8\%\\
		STAMP  & 89.4\%& 89.5\%& 77.1\%& 88.9\%& 89.7\%& 76.0\%\\
		STAMP+RL& 89.6\%&  89.7\%& 77.3\%&  90.0\%&  89.9\%& 76.3\%\\
		\hline
	\end{tabular}
	\caption{Fine-grained accuracies on the WikiSQL dev and test sets. Accuracy (Acc$_{lf}$) is evaluated on SELECT column (Acc$_{sel}$) , SELECT aggregator (Acc$_{agg}$), and WHERE clause (Acc$_{where}$), respectively. }
	\label{table:fine-grained-results}
\end{table*}

{Firstly}, since SQL queries in WikiSQL \mbox{consists} of SELECT column, SELECT aggregator, and WHERE clause, we report the results with regard to more fine-grained evaluation metrics over these aspects.
Results are given in Table \ref{table:fine-grained-results}, in which the numbers of Seq2SQL and SQLNet are reported in \cite{xu2017sqlnet}.
We can see that the main improvement of STAMP comes from the WHERE clause, which is also the key challenge of the WikiSQL dataset.
This is consistent with our primary intuition on improving the prediction of WHERE column and WHERE value.
The accuracies of \mbox{STAMP} on SELECT column and SELECT aggregator are not as high as SQLNet.
The main reason is that these two approaches train the \mbox{SELECT} clause separately while STAMP learns all these components in a unified paradigm.

\vspace{-0.3cm}
\subsection{Model Analysis: Difficulty Analysis}
We study the performance of STAMP on different portions of the test set according to the difficulties of examples.
We compare between Aug.PntNet (re-implemented by us) and STAMP.
In this work, the difficulty of an example is reflected by the number of WHERE columns.

\begin{table}[h]
	\centering
	\begin{tabular}{p{2cm}lcc}
		\hline
		Method & \#where & Dev & Test\\
		\hline
		\multirow{3}{*}{Aug.PntNet} &  $=$ 1  & 63.4\%& 63.8\% \\
		& $=$ 2   & 51.0\%& 51.8\%\\
		& $\geq$ 3  & 38.5\%& 38.1\%\\
		\hline
		\multirow{3}{*}{STAMP} &  $=$ 1 & 80.9\%& 80.2\%\\
		&$=$ 2  &  65.1\%& 65.4\%\\
		& $\geq$ 3 &  44.1\%& 48.2\%\\
		\hline
	\end{tabular}
	\caption{Execution accuracy (Acc$_{ex}$) on different groups of WikiSQL dev and test sets. }
	\label{table:difficulty}
\end{table}
From Table \ref{table:difficulty}, we can see that STAMP outperforms Aug.PntNet in all these groups.
The accuracy decreases with the increase of the number of WHERE conditions.


\subsection{Model Analysis: Executable Analysis}
We study the percentage of executable SQL queries in the generated results.
As shown in Table \ref{table:executable}, STAMP significantly outperforms Aug.PntNet.
Almost all the results of STAMP are executable.
This is because STAMP avoids generating incomplete column names or cells, and guarantees the correlation between WHERE conditions and WHERE values in the table.

\begin{table}[h]
	\centering
	\begin{tabular}{l|c|c}
		\hline
		& Dev& Test\\
		\hline
		Aug.PntNet &  77.9\%& 78.7\%\\
		STAMP  & 99.9\%& 99.9\%\\
		\hline
	\end{tabular}
	\caption{Percentage of the executable SQL queries on WikiSQL dev and test sets. }
	\label{table:executable}
\end{table}

\subsection{Model Analysis: Case Study}
\begin{figure*}[t]
	\centering
	\includegraphics[width=.92\textwidth]{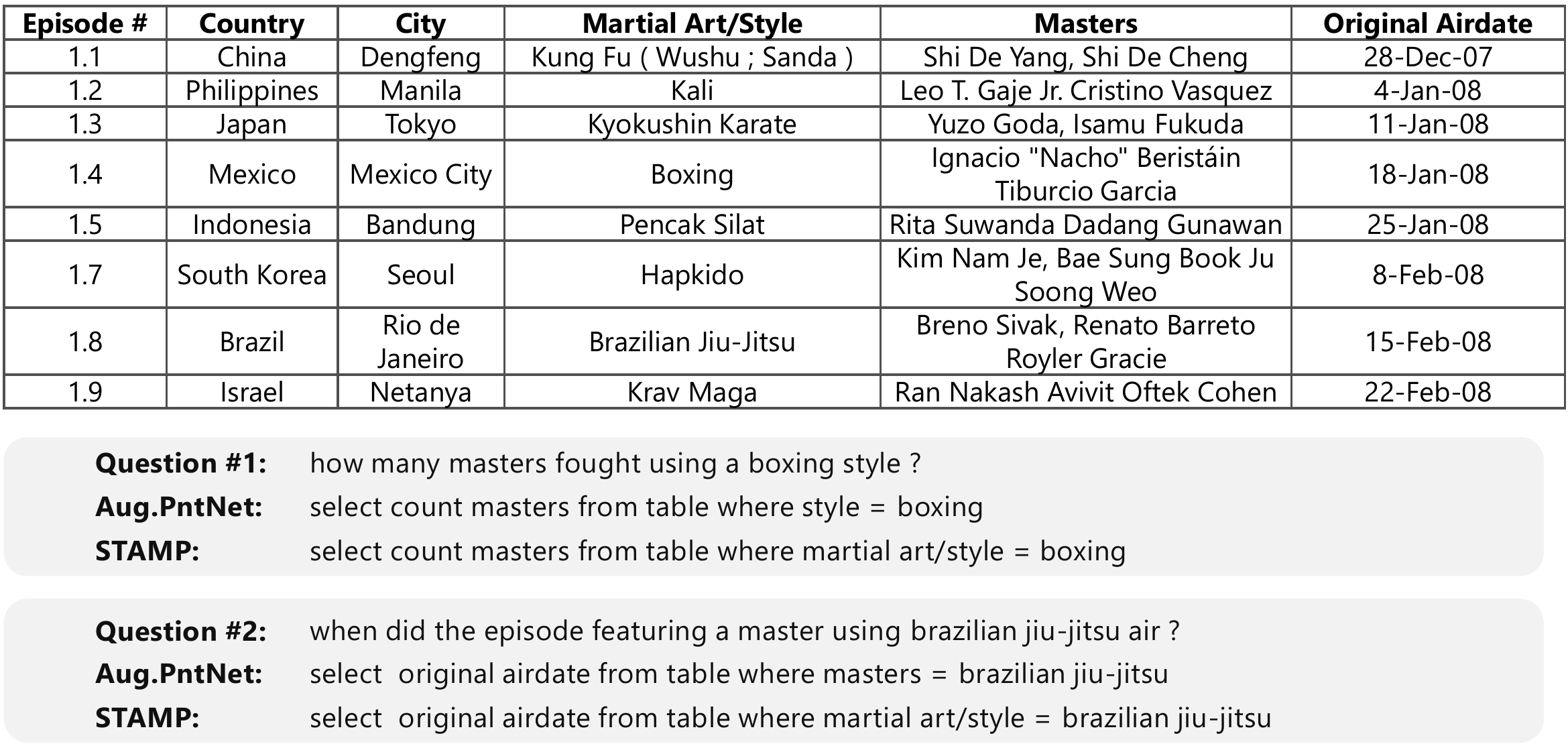}
	\caption{Case study on the dev set between Aug.PntNet and STAMP. These two questions are based on the same table. Each question is followed by the generated SQL queries from the two approaches.}
	\label{fig:example}
\end{figure*}

We give a case study to illustrate the generated results by STAMP, with a comparison to Aug.PntNet.
Results are given in Figure \ref{fig:example}.
In the first example, Aug.PntNet generates incomplete column name (``\textit{style}''), which is addressed in STAMP through replicating an entire column name.
In the second example, the WHERE value (``\textit{brazilian jiu-jitsu}'') does not belong to the generated WHERE column  ({``\textit{Masters}''}) in Aug.PntNet. This problem is avoided in STAMP through incorporating the table content.

\subsection{Error Analysis}
We conduct error analysis on the dev set of WikiSQL to show the limitation of the STAMP model and where is the room for making further improvements.
We analyze the 2,302 examples which are executed to wrong answers by the STAMP model, and find that 33.6\% of them have wrong SELECT columns, 15.7\%  of them have a different number of conditions in the WHERE clause, and 53.7\% of them have a different WHERE column set compared to the ground truth.
Afterwards, we analyze a portion of randomly sampled dissatisfied examples.
Consistent with the qualitative results, most problems come from column prediction, including both SELECT clause and WHERE clause. Even though the overall accuracy of the SELECT column prediction is about 90\% and we also use cell information to enhance the column representation, this semantic gap is still the main bottleneck.
Extracting and incorporating various expressions for a table column (i.e. relation in a relational database) might be a potential way to mitigate this problem.
Compared to column prediction, the quality of cell prediction is much better because cell content typically (partially) appears in the question.

\subsection{Transfers to WikiTableQuestions}
WikiTableQuestions \cite{pasupat-liang:2015:ACL-IJCNLP} is a widely used dataset for semantic parsing.
To further test the performance of our approach, we conduct an additional transfer learning \mbox{experiment}.
Firstly, we directly apply the STAMP model trained on WikiSQL to WikiTableQuestions, which is an unsupervised learning setting for the WikiTableQuestions dataset.
Results show that the test accuracy of STAMP in this setting is 14.5\%, which has a big gap between best systems on  WikiTableQuestions, where \newcite{zhang-pasupat-liang:2017:EMNLP2017} and \newcite{krishnamurthy-dasigi-gardner:2017:EMNLP2017} yield 43.3\% and 43.7\%, respectively.
Furthermore, we apply the learnt STAMP model to generate SQL queries on natural language questions from WikiTableQuestions, and regard the generated SQL queries which could be executed to correct answers as additional pseudo question-SQL pairs. 
In this way, the STAMP model learnt from a combination of WikiSQL and pseudo question-SQL pairs could achieve 21.0\% on the test set.
We find that this big gap is caused by the difference between the two datasets.
Among 8 types of questions in WikiTableQuestions, half of them including \{``\textit{Union}'', ``\textit{Intersection}'', ``\textit{Reverse}'', ``\textit{Arithmetic}''\} are not covered in the WikiSQL dataset.
It is an interesting direction to leverage algorithms developed from two datasets to improve one another. 

\subsection{Discussion}
Compared to slot-filling based models that restrict target SQL queries to fixed forms of ``select-aggregator-where'', our model is less tailored. We believe that it is easy to expand our model to generate nested SQL queries or JOIN clauses, which could also be easily trained with back-propagation if enough training instances of these SQL types are available.  For example, we could incorporate a hierarchical ``value'' channel to handle nest queries. Let us suppose our decoder works horizontally that next generated token is at the right hand of the current token. Inspired by chunk-based decoder for neural machine translation \cite{ishiwatari-EtAl:2017:Long}, we could increase the depth of the ``value'' channel to generates tokens of a nested WHERE value along the vertical axis. During inference, an addition gating function might be necessary to determine whether to generate a nested query, followed by the generation of WHERE value. 
An intuitive way that extends our model to handle JOIN clauses is to add the 4th channel, which predicts a table from a collection of tables. Therefore, the decoder should learn to select one of the four channels at each time step. Accordingly, we need to add ``from'' as a new SQL keyword in order to generate SQL queries including ``from xxxTable''.

In terms of the syntax of SQL, the grammar we used in this work could be regarded as shallow syntax, such as three channels and column-cell relation. We do not use deep syntax, such as the sketch of SQL language utilized in some slot-filling models, because incorporating them would make the model clumpy. Instead, we let the model to learn the sequential and compositional relations of SQL queries automatically from data. Empirical results show that our model learns these patterns well.

\section{Conclusion and Future Work}
In this work, we develop STAMP, a Syntax- and Table- Aware seMantic Parser that automatically maps natural language questions to SQL queries, which could be executed on web table or relational dataset to get the answer.
STAMP has three channels, and it \mbox{learns} to switch to which channel at each time step.
\mbox{STAMP} considers cell information and the relation between cell and column name in the generation process.
Experiments are conducted on the WikiSQL dataset. Results show that STAMP achieves the new state-of-the-art performance on WikiSQL.
We conduct extensive experiment analysis to show
advantages and limitations of our approach, and where is the room for others to make further improvements.

SQL language has more complicated queries than the cases included in the WikiSQL dataset, including (1) querying over multiple relational databases, (2) nested SQL query as condition value, (3) more operations such as ``\textit{group by}'' and ``\textit{order by}'', etc.
In this work, the STAMP model is not designed for the first and second cases, but it could be easily adapted to the third case through incorporating additional SQL keywords and of course the learning of which requires dataset of the same type.
In the future, we plan to improve the accuracy of the column prediction \mbox{component}.
We also plan to build a large-scale dataset that considers more sophisticated SQL queries.

\bibliography{naaclhlt2018}
\bibliographystyle{acl_natbib}

\appendix

%

\end{document}